\documentclass[conference]{IEEEtran}
\IEEEoverridecommandlockouts
\usepackage{cite}
\usepackage{amsmath,amssymb,amsfonts}
\usepackage{algorithmic}
\usepackage{bbm}
\usepackage{graphicx}
\usepackage{textcomp}
\usepackage{xcolor}
\def\BibTeX{{\rm B\kern-.05em{\sc i\kern-.025em b}\kern-.08em
    T\kern-.1667em\lower.7ex\hbox{E}\kern-.125emX}}

\begin{document}

\title{FedMetaMed: Federated Meta-Learning for Personalized Medication in Distributed Healthcare Systems}

% \and
% \and
\author{\IEEEauthorblockN{ Jiechao Gao* \thanks{* Corresponding author: Jiechao Gao}}
\IEEEauthorblockA{\textit{Department of Computer Science} \\
\textit{University of Virginia}\\
Charlottesville, VA, USA\\
jg5ycn@virginia.edu}
\and
\IEEEauthorblockN{Yuangang Li}
\IEEEauthorblockA{\textit{Thomas Lord Department of Computer Science} \\
\textit{University of Southern California}\\
Los Angeles, CA, USA \\
yuangang@usc.edu}

}

\maketitle

\begin{abstract}

Personalized medication aims to tailor healthcare to individual patient characteristics. However, the heterogeneity of patient data across healthcare systems presents significant challenges to achieving accurate and effective personalized treatments. Ethical concerns further complicate the aggregation of large volumes of data from diverse institutions. Federated Learning (FL) offers a promising decentralized solution by enabling collaborative model training through the exchange of client models rather than raw data, thus preserving privacy. However, existing FL methods often suffer from retrogression during server aggregation, leading to a decline in model performance in real-world medical FL settings. To address data variability in distributed healthcare systems, we introduce Federated Meta-Learning for Personalized Medication (FedMetaMed), which combines federated learning and meta-learning to create models that adapt to diverse patient data across healthcare systems. The FedMetaMed framework aims to produce superior personalized models for individual clients by addressing these limitations. Specifically, we introduce Cumulative Fourier Aggregation (CFA) at the server to improve stability and effectiveness in global knowledge aggregation. CFA achieves this by gradually integrating client models from low to high frequencies. At the client level, we implement a Collaborative Transfer Optimization (CTO) strategy with a three-step process—Retrieve, Reciprocate, and Refine—to enhance the personalized local model through seamless global knowledge transfer. Experiments on real-world medical imaging datasets demonstrate that FedMetaMed outperforms state-of-the-art FL methods, showing superior generalization even on out-of-distribution cohorts.

\end{abstract}


\section{Introduction} 
To fully unlock the potential of precision medicine, it is essential to establish interoperability and broad accessibility of extensive medical datasets for researchers \cite{1}. However, the current landscape poses significant challenges, as medical data are fragmented across numerous institutions, making centralized access and aggregation nearly impossible \cite{tang2023srda}. These challenges are not primarily technical—transferring heterogeneous data across organizations is feasible—but are instead rooted in legal and regulatory barriers \cite{3}. Transferring patient-level data beyond healthcare providers is a complex, time-consuming process due to stringent legal and regulatory requirements. Given the scarcity and biased distribution of medical data, there is an increasing call for collaborative data sharing among medical institutions to enhance model performance in critical tasks like disease diagnosis\cite{ai_dt}. Federated learning (FL) has emerged as a promising solution, enabling the use of distributed data sources while preserving privacy. In FL, instead of transmitting raw data, clients—represented by various devices or locations—share gradients of their locally trained model parameters with a central server. The server then performs a weighted aggregation of these gradients to update the global model, which is subsequently distributed back to all clients for further training iterations \cite{5,liu2024fedbcgd}.

% FL achieves this without compromising sensitive health data, as it only shares model parameters, ensuring privacy and security by keeping the raw data within the clients' devices .
While federated learning leverages data from multiple clients, its performance struggles with heterogeneous datasets~\cite{gao2022pfed}, as real-world data from different clients often focus on different tasks \cite{gao2022residential,6, gao2023pfdrl}. Federated learning also faces scalability issues, as increased participants lead to communication bottlenecks. Additionally, privacy concerns are heightened due to the sensitive nature of medical data, requiring robust protections during collaborative model training. Balancing collaboration effectiveness with strict privacy measures is a significant challenge in healthcare and addressing these scalability and privacy issues is essential for federated learning’s successful integration into medical research and diagnostics, emphasizing the need for continued innovation in this space.
In real-world medical Federated Learning (FL) scenarios, variations in client data from different imaging devices, protocols, and regional disease characteristics create disparities between local models \cite{4}. This causes performance drops after each communication cycle, as aggregated models lose prior knowledge and require re-adaptation during local training, disrupting effective knowledge sharing \cite{8}. The decline in FL performance in these settings stems from two main factors. First, severe data diversity makes element-wise parameter averaging ineffective because parameters from different clients may represent different semantic patterns \cite{Gao2024FedLDR}. Representing parameters in the frequency domain allows for better alignment and selective aggregation of frequency bands. Second, replacing local models with the aggregated server model erases the knowledge acquired by local models, hampering optimization in subsequent training iterations.
% In real-world medical scenarios involving Federated Learning, the diversity of data among clients, due to variations in imaging devices, protocols, and regional disease characteristics, poses a more significant challenge than seen in previous studies. This leads to substantial disparities between local models \cite{4}. Existing FL techniques encounter a sudden drop in performance following each communication cycle between the server and clients. Aggregated models lose their previous knowledge and require re-adaptation during local training, impeding both client training and server-side knowledge sharing \cite{8}.

% The decline in real-world Federated Learning performance may be attributed to two primary reasons. Firstly, when severe data diversity exists among clients, averaging parameters element-wise becomes unreasonable as these parameters from different clients may represent distinct semantic patterns at the same location. In contrast, representing parameters in the frequency domain ensures alignment along the frequency dimension and permits the selection of frequency bands for aggregation. Secondly, replacing the local models with the aggregated server model during communication erases the knowledge acquired by the local models, thereby undermining optimization in subsequent iterations.
This study introduces a novel approach to address the challenges of personalized medicine in distributed healthcare systems. We propose FedMetaMed, a personalized federated meta-learning framework that provides tailored models for each client. On the server side, we implement Cumulative Fourier Aggregation (CFA), which uses Fast Fourier Transform (FFT) and inverse FFT (IFFT) to convert client parameters to the frequency domain. Low-frequency components are averaged while high-frequency components are retained, with the frequency threshold gradually increasing throughout training to align client knowledge with network preferences. Notably, this is the first work to apply frequency-domain parameter aggregation in Federated Learning.
% To address these issues and the challenges in Personalized Medicine within Distributed Healthcare Systems, a novel approach is introduced in this study. A personalized federated meta-learning framework is proposed known as FedMetaMed from both server and client perspectives, aiming to provide a tailored model for each client. A technique called Cumulative Fourier Aggregation (CFA) is devised for server-side integration of client models. This method employs Fast Fourier Transform (FFT) and inverse FFT (IFFT) to convert client parameters to the frequency domain. Low-frequency components of client parameters are averaged while retaining high-frequency components. The frequency threshold of shared components is gradually increased during Federated Learning, ensuring integration of client knowledge aligned with network learning preferences. Notably, this work represents the first endeavor to implement parameter aggregation in the frequency domain for Federated Learning.
On the client side, instead of replacing local models entirely, we propose Collaborative Transfer Optimization (CTO). This enables a client model to receive updates from the server model while maintaining its personalized local model without contamination. Since client models tend to experience performance drops after each communication, the CTO method encompasses three steps: Retrieve, Reciprocate, and Refine. These steps collectively restore the local model's prior state and facilitate the transfer of comprehensive knowledge to enhance the personalized local model.

Extensive experiments on real-world medical Federated Learning datasets confirm the effectiveness and generalization capability of the proposed FedMetaMed framework, which outperforms existing state-of-the-art FL methods. We analyze FedMetaMed’s relationship with the original MetaFed algorithm, discuss key considerations for its implementation, and establish its convergence properties in minimizing loss functions. Notably, we examine how data heterogeneity and distribution similarity among users, measured by distribution distances, influence the convergence behavior of FedMetaMed.

\begin{figure*}[t]
\centering
      \includegraphics[scale=0.5]{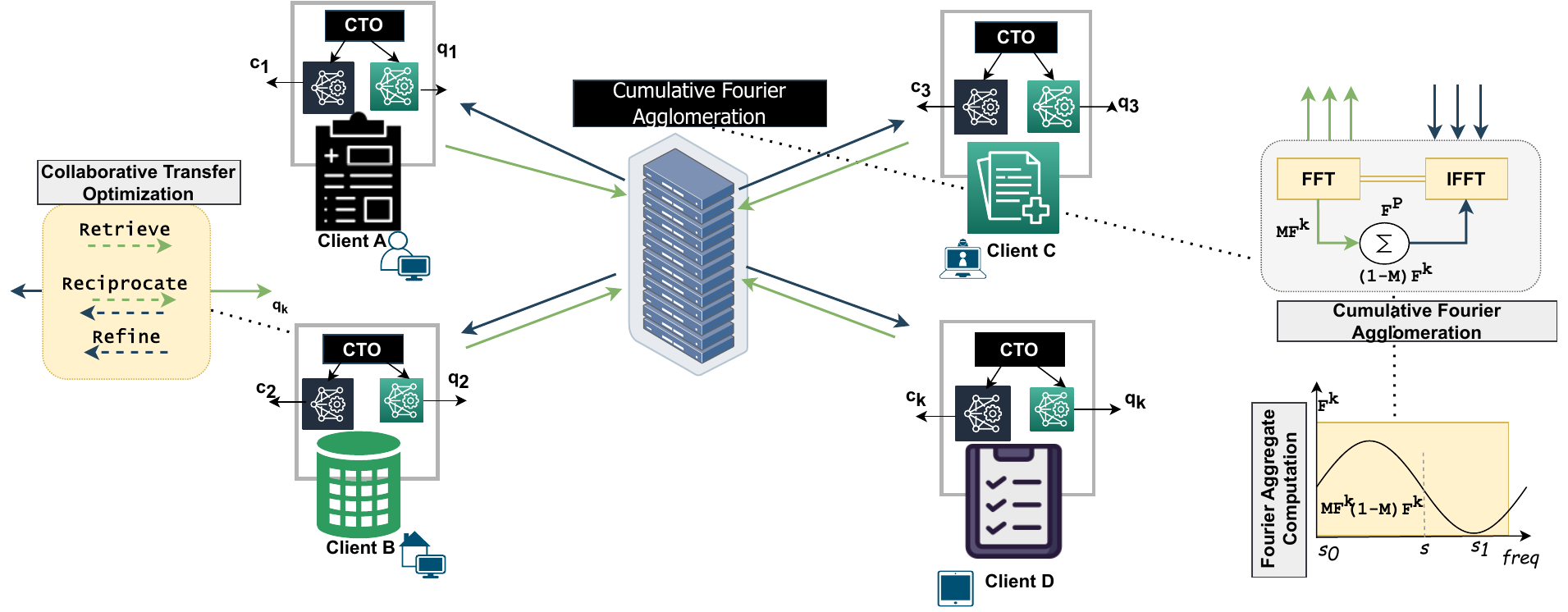}
\caption{The diagram illustrates FedMetaMed. The server utilizes Cumulative Fourier Aggregation (CFA) to amalgamate comprehensive knowledge in the frequency domain. Meanwhile, each client employs Collaborative Transfer Optimization (CTO) to enhance its local personalized model, while minimizing the disruption caused by communication.}\label{figsystem}
\end{figure*}

\section{Related work}
Deep learning has shown exceptional potential in healthcare, from activity recognition to personalized health models. However, centralized data collection is often impractical due to privacy concerns, as data is generated by numerous individuals and organizations reluctant to share sensitive information. FedAvg \cite{20} has emerged as a solution, where locally trained models are averaged across clients, treating each as an independent and heterogeneous entity, while enabling collaborative healthcare applications such as identifying patients with similar clinical profiles.
% Deep learning has showcased its remarkable capabilities across various fields, particularly in healthcare. From recognizing everyday human activities to personalized healthcare models, deep learning has proven its worth. In the healthcare domain, obtaining centralized data is often an unattainable goal, as multiple individuals or organizations generate vast amounts of data while being hesitant to share their private information. Numerous research studies and innovative projects have embraced federated approaches to tackle various healthcare challenges.
%  Federated methods have been employed to identify patients with similar clinical characteristics, predict hospitalizations based on disease severity, and make predictions related to ICU stay duration and mortality. Google's introduction of FedAvg \cite{20} demonstrated the concept that averaging the models trained locally on various clients extends the idea of updating batch gradients when considering clients as independent and heterogeneous entities. This approach helps in minimizing the communication overhead.

To address varying data distributions among clients, FedProx \cite{30} introduced a proximal term to align local updates with the initial global model. Researchers like Brismi et al. \cite{15} have incorporated adaptation techniques into federated models to improve performance, while approaches like FedBN \cite{23} and SiloBN~\cite{andreux2020siloed} tackle feature shifts by preserving local batch normalization parameters, allowing each client to handle unique data distributions effectively. Personalized FL, which lets clients choose between the aggregated server model and intermediate client models during training, is particularly useful for medical applications.

Frameworks such as FedHealth \cite{22} for wearable healthcare devices and FED-ROD \cite{24} for balancing generalization and personalization demonstrate the versatility of federated methods in healthcare. Personalized FL approaches, though promising, often face limitations due to communication costs and server dependencies. Otoum et al. \cite{27} analyzed data from various hospitals to identify patient phenotypes while keeping data siloed by hospital. Results showed that federated learning, which preserves data locality, achieved comparable accuracy and phenotype insights to a centralized model, thus safeguarding privacy \cite{26}. Chen et al. \cite{36} presented MetaFed, a framework that enables trustworthy Federated Learning (FL) across diverse federations without a central server. Using Cyclic Knowledge Distillation, MetaFed creates personalized models through two phases: common knowledge accumulation and personalized model refinement.

Although several federated learning methods have been used in initial studies with clinical data, they tend to neglect the typical challenges seen in analyzing biomedical datasets. Medical data has distinct characteristics due to ethical, legal, and social factors, along with imbalances in phenotype prevalence and cohort sizes.

\section{Methodology}
% Federated learning harnesses the collective computational power of all participants while preserving the privacy of user data. It empowers model training by tapping into a diverse array of clients that offer distinct datasets. Nevertheless, despite the advantages of federated learning in aggregating more data for a shared objective, its efficacy diminishes when confronted with heterogeneous datasets originating from disparate clients.

% To confront this challenge, it is imperative to embrace the exquisite approach of Meta Learning, which endeavors to cultivate an adaptable model capable of navigating diverse tasks effortlessly. By incorporating federated learning, we can train the model utilizing datasets sourced from various clients, while meta learning empowers us to forge an adaptive model through heterogeneous datasets. Thus, our objective is to delve into the profound essence of meta learning, harnessing its core principles to architect an adaptive federated learning algorithm that propels successful training of our model amidst the vast array of heterogeneous medical datasets.

% In Federated learning, model parameters are directly selected for integration, whereas in federated meta learning, losses from multiple training tasks are integrated. Our objective is to design an adaptive federated learning approach based on the fundamental concept of meta learning. By combining these two algorithms, we introduce FedMetaMed.

Federated learning leverages the combined computational power of diverse clients while preserving user privacy, enabling model training across varied datasets. However, its effectiveness decreases when faced with heterogeneous data from different clients. To address this, we propose combining federated learning with meta-learning to create an adaptable model capable of handling diverse tasks. Meta-learning allows us to train a model that adapts to heterogeneous data, while federated learning facilitates training across decentralized datasets. By merging these principles, we introduce FedMetaMed—an adaptive federated learning framework designed for success with varied medical datasets. In standard federated learning, model parameters are aggregated, whereas in federated meta-learning, losses from multiple tasks are integrated. FedMetaMed utilizes this adaptive approach to optimize training across heterogeneous client data.

In our customized regression-resistant framework, our objective is to collectively create a personalized model $p$ with superior performance for each of the $N$ models. These individual models $x_n$ (for $n$ ranging from 1 to $N$) share the same network structure, allowing them to benefit from server aggregation, which is similar to previous federated learning approaches. For the $n^{th}$ model, the personalized model $x_n$ is trained locally with private data for $L$ epochs and then uploaded to the server. The server gathers the client models and combines them into server model, incorporating the high-frequency components unique to each client, using our FedMetaMed technique. Subsequently, these server models are transmitted back to the respective clients as temporary models, enabling the exhaustive knowledge transfer through our FedMetaMed mechanism. These steps are repeated until the local training reaches $M$ epochs. Our model can be elegantly expressed as a function \begin{math}k_\sigma \end{math}, where \begin{math}
     \sigma
 \end{math} represents a vector of parameters. For every individual training task (i), we are required to calculate the losses \begin{math}
    Ui(k_\sigma)
\end{math} using samples specifically designated for that task. These losses are then combined in a cohesive manner. Notably, our primary model remains unaltered throughout this meta-training process. Subsequently, the integrated loss is utilized to perform a backward pass and modify our main model's parameters.

% The learning rates for meta training models and our main model are denoted as $\alpha$  and $\beta$ respectively. The parameters for the $k^{th}$ training task are represented as  $\theta^k$, while  $\theta$ signifies the parameters for the main model. In a similar manner, each training task has the option to perform multiple epochs of meta gradient descent before integrating the loss. Furthermore, in equation, we observe that the main model parameters $\theta$ are trained by optimizing the integrated loss, which relies on different meta training model parameters $\theta^k$.

The methodology in the following provides an overview of our personalized regression-resistant framework. Figure \ref{figsystem}
shows a high level overview of the proposed FedMetaMed Architecture. It  depicts a customized and diverse federated meta-learning structure. The server employs Cumulative Fourier Aggregation (CFA) to combine extensive information in the frequency domain. In the meantime, every client utilizes Collaborative Transfer Optimization (CTO) to improve its individualized local model, while reducing the impact of communication disturbances.

\subsection{Cumulative Fourier Agglomeration}
% The previous studies adopted a method of aggregating the exhaustive knowledge from various clients by directly averaging the parameters of local models. However, this approach, which involves a coarse operation in the parameter space, significantly harms the performance of the models on individual clients. In order to mitigate the regression caused by aggregation, we introduce a new technique called Cumulative Fourier Agglomeration (CFA) that allows for stable integration of client models in the frequency domain.

Our CFA (Cumulative Frequency Agglomeration) is motivated by the observation that the low-frequency components of network parameters play a crucial role in determining the network's capability. To leverage this insight, our approach combines the relatively low-frequency components of data points from different clients to share knowledge, while retaining the high-frequency components that may contain client-specific information.

Specifically, in the case of a convolutional layer belonging to the $n^{th}$ client model, we begin by reshaping its parameter tensor \begin{math}
    w_n \in \mathbb{R}^{A x B x c1 x c2}
\end{math} 
 into a 2-D matrix $w'_n \in \mathbb{R}^{c1A \times c2B}$ 

where $A$ and $B$ denote the output and input channels respectively, and $c1$ and $c2$ represent the spatial dimensions of the kernel. Next, we employ the Fourier transform to obtain the amplitude map $F^b$
and the phase map \begin{math} F^q  \end{math}, resulting in $F = F^b {e^jF}^q $
The procedure can be summarized as follows:

\begin{equation}
    F(w'_n)(a,b)=\sum_{(i,j)} (w'_n)(i,j)e^{-k2\pi(\frac{i}{c_1A}a+ \frac{j}{c_2B}b}), q^2=-1
\end{equation}

We can efficiently implement the given process by utilizing the fast fourier transform. In order to gather the low-frequency elements for agglomeration, we utilize a low-frequency mask A. This mask mainly consists of zero values, except for the main region:

\begin{equation}
    N(a, b) = \mathbbm{1}_{(a,b) \in [-sc_1A:sc_1A,-sc_2B:sc_2B]}
\end{equation}

The low-frequency threshold, denoted as s \begin{math}
    \in (0 , 0.5)
\end{math} defines the range, and the center of \begin{math}
    w'_n
\end{math}
is positioned at the coordinates $(0, 0)$. By taking the average of the low-frequency elements across all clients, the combined frequency components for the n-th client are computed as:
\begin{equation}
    F^{k}(w'_n) = (1 - M)  \cdot F^k(w'^n) + \frac{1}{N} \sum_{n=1}^{N} M F^k(w'_n),
\end{equation}

% \begin{equation}
%     F^{k}(w'_n) = (1 - M) \cdot F^k(w'_n) + \frac{1}{N} \sum_{n=1}^{N} M \odot F^k(w'_n),
% \end{equation}

In light of the fact that networks are typically trained to acquire low-frequency knowledge before high-frequency information, we propose a cumulative approach for implementing our CFA. This involves gradually increasing the value of \begin{math}
    s = s_0 + \frac{(s_1 - s_0)}{T} * t
\end{math} during FL training, where \begin{math}
    s_0
\end{math} and \begin{math}
    s_1
\end{math} represents the initial and final low-frequency thresholds. 

% Additionally, the element-wise multiplication is denoted by \cdot.

After performing the inverse Fourier transform $F^{-1}$ on the amplitude and phase maps to revert them back to parameters, we achieve the aggregated parameters of the $n^{th}$ client as \begin{math}
    w'_n = F^{-1}([F^k(w_n), F^p(w_n)]).
\end{math}
Additionally, the identical CFA can be employed on the fully connected layers of client models, eliminating the need to reshape parameters.
\subsection{Collaborative Transfer Optimization}
% Our CFA can mitigate the setback caused by clustering on the server, but when we directly update client models with the combined server parameters, it erases the acquired local knowledge and further worsens the optimization process in the next iteration. To address this issue, we introduce Collaborative Transfer Optimization (CTO), which tactfully merges exhaustive knowledge with local expertise instead of simply replacing it. Alongside an individualized local model q, each client also includes a client model c to receive the combined parameters from the server. The FedMetaMed CTO consists of three steps: Retrieve, Reciprocate, and Refine, to seamlessly transfer the exhaustive knowledge from the client c to the personalized local model q.

Our CFA can mitigate the setback caused by clustering on the server, but when we directly update client models with the combined server parameters, it erases the acquired local knowledge and further worsens the optimization process in the next iteration. To address this issue, we introduce Collaborative Transfer Optimization (CTO), which tactfully merges exhaustive knowledge with local expertise instead of simply replacing it. Alongside an individualized local model $q$, each client also includes $a$ client model $c$ to receive the combined parameters from the server. The FedMetaMed CTO consists of three steps: Retrieve, Reciprocate, and Refine, to seamlessly transfer the exhaustive knowledge from the client $c$ to the personalized local model $q$.

\subsection{Retrieve}
Upon receiving the aggregated model $s$ from the server, the client $c$ experiences a significant decline in performance caused by the aforementioned retrogress issue. Consequently, our initial approach involves utilizing the personalized local model $q$ as a teacher to restore the client model $c$ using its local knowledge. During this phase, the personalized local model $q$ is guided by the cross entropy loss, while the client model $c$ is optimized using a joint loss function \begin{math}
  L_c  
\end{math}
\begin{equation}\label{eq4}
    L_c = L_{en} + \sum_{j=1}^{M}q(y_i) \log \frac{q(y_i)}{r(y_i)}
\end{equation}
In the given expression,  \begin{math}
 y_i   
\end{math} represents the $i^{th}$ training sample, and M denotes the total number of training samples on the current client. The loss function used is \begin{math}
    L_{en}
\end{math}, which corresponds to cross-entropy. The terms \begin{math}
    q(y_i)
\end{math} and \begin{math}
    r(y_i)
\end{math}  refer to the posterior probabilities of the client $c$ and the personalized local model $q$, respectively. The second part of Equation \ref{eq4} represents the Kullback-Leibler (KL) divergence, which aids the client $c$ in swiftly recovering adaptability while enhancing performance. This particular step is crucial to ensure that the deputy model does not adversely impact the personalized local model in subsequent steps.

\subsection{Reciprocate}
When the client $c$ achieves a performance level that is similar to the model $q$, indicated by  $\Phi $ val(c) $\geq \lambda \Phi$ val(q)  where  $\Phi$val represents a specific performance metric on the validation set, we initiate mutual learning. This process involves reciprocating exhaustive information between the client $c$ and the personalized local model $q$. The objective of mutual learning is to interchange both overall knowledge and local knowledge between the two models. The client model $c$ is supervised using Equation \ref{eq5} while the personalized local model $q$ is trained using the following loss function, denoted as $L_q$.

\begin{equation} \label{eq5}
    L_q = L_{en} +\sum_{i=1}^{N} \frac{r(y_i)log r(y_i)}{q(y_i)}
\end{equation}
The client $c$ facilitates the transfer of exhaustive knowledge from the server to the personalized local model $q$, as expressed in the second term of Equation. This process enhances the generalization capabilities of all clients by promoting knowledge interchange.

\subsection{Refine}

Ultimately, when the client's performance, represented by $\phi$ $c$, closely matches the personalized local model, represented by $\phi$ $q$, and satisfies the condition $\phi$(c) $\geq$ $\lambda$2$\phi$(q), where 0 $ \leq $ $\lambda$  1 $\leq$
$\lambda$ 2 $\leq$ 1, the client model $c$ assumes the role of a teacher, guiding the personalized local model $q$ further using $L_q$ from the Equation \ref{eq5}. This process ensures that the maximum amount of global knowledge is transferred to the personalized local model.

\subsection{Privacy Implications \& Preservation}
The personalized federated meta-learning model for distributed systems leverages Cumulative Fourier Aggregation (CFA) at the server to enhance global knowledge integration by gradually combining client models across the frequency spectrum. A dedicated client model receives the aggregated server model, improving stability and effectiveness. To address privacy concerns, the model incorporates Collaborative Transfer Optimization (CTO) on the client side, using three steps—Retrieve, Reciprocate, and Refine—to smoothly transfer global knowledge to the personalized local model. Compared to existing approaches, this model stands out by combining Fourier Aggregation, Transfer Optimization, and multi-step refinement, offering a more robust solution to privacy preservation in federated learning.

\section{Experiments}
\subsection{Dataset Detail}
\begin{figure}[t]
\centering
      \includegraphics[width=5.5cm]{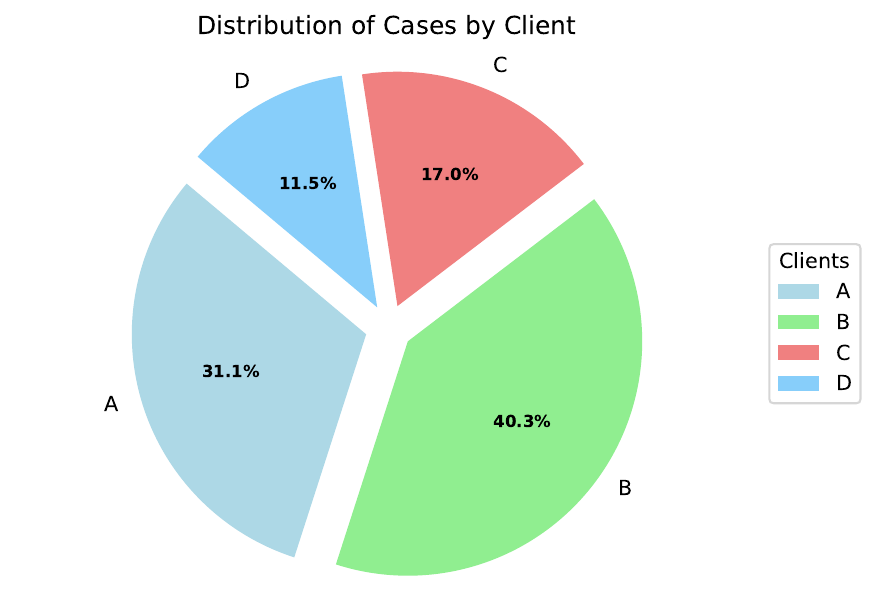}
\caption{Dataset distribution percentage across clients.}
\label{figuredata}
\end{figure}

\paragraph{\normalfont\textbf{HAMK}}
In this research, we used the HAMK dataset, an open-source collection of 10,015 dermoscopic images categorized into seven classes. Like many medical imaging datasets, HAMK is highly imbalanced, with 67\% of images labeled as nevi, 11\% as melanoma, and the remaining 22\% spread across five other classes. Ground truth labels in HAMK are based on histopathology in over 50\% of cases, with the rest determined by follow-up exams, expert consensus, or in-vivo confocal microscopy. We selected HAMK for evaluation due to its unbalanced and heterogeneous nature. Table \ref{tabledata} details the sample counts for each class, and Figure \ref{figuredata} shows the client-wise distribution as a pie chart.
% In this research, we have utilized HAMK dataset, an openly accessible dataset containing 10,015 dermoscopic images categorized into seven classes. Similar to many medical imaging datasets, HAMK exhibits a significant class imbalance, with 67\% of the data representing nevi, 11\% melanoma, and the remaining 22\% corresponding to the other five classes. The HAMK dataset's ground truth labels rely on histopathology in more than 50\% of the cases, with the remaining cases determined through follow-up examination, expert consensus, or in-vivo confocal microscopy. We selected HAMK to evaluate our method because it exhibits an unbalanced and highly heterogeneous nature.  Table \ref{tabledata} shows sample count details of all heterogeneous records and their distribution with respect to clients is represented in the form of pie chart in Figure \ref{figuredata}.

\paragraph{\normalfont\textbf{MSK}}
The MSK dataset is from the Memorial Sloan-Kettering Cancer Center and used in ISIC lesion recognition challenges. ISIC has compiled over 20,000 dermoscopy images from top clinical centers worldwide, captured using various devices. This dataset served as the basis for the 2016 benchmark challenge on dermoscopic image analysis.

\subsection{Implementation}

We have employed a customized framework along with cutting-edge Federated Learning (FL) techniques utilizing the VGG-16 BN architecture in the PyTorch framework. At the client side, local training is carried out, and the networks are optimized using Stochastic Gradient Descent (SGD) with a batch size of 20. The learning rate is initialized to $3 x 10^-3$ and is halved after every 30 epochs. Communication among clients occurs every E = 10 epochs during local training, continuing until the total training reaches T = 300 epochs.

 To enhance the Federated Averaging (CFA) method, we have set the frequency thresholds $s_0$ and $s_1$ to 0.26 and 0.55, respectively. Furthermore, in the Collaborative Transfer optimization (CTO), the performance  $\lambda 1$ and $\lambda 2$ are set to 0.6 and 0.8, respectively. To evaluate the performance of both client and personalized models, we adopt the F1 score as the metric $\phi val$.

\begin{table}[t]
\caption{Samples per client along with the corresponding label, all client datasets originate from HAMK and MSK.}  \label{table1} 
\centering
\begin{tabular}{lllll}

\hline
% \rowcolor{Blue}
\textbf{Client} & \textbf{Angioma} & \textbf{Benign}  & \textbf{Melanoma} & \textbf{Total} \\ \hline
\textbf{A}      & 1832  & 475              & 680      & 2987  \\ 
% \rowcolor{LightBlue}
\textbf{B}      & 3720  & 124              & 24       & 3868  \\ 
\textbf{C}      & 803   & 490              & 342      & 1635  \\ 
% \rowcolor{LightBlue}
\textbf{D}      & 1372  & 254              & 374      & 2000 \\ \hline
\end{tabular}
\label{tabledata}
\end{table}

Model selection is done by employing the hold-out validation set on each client to determine the best-performing model. For data augmentation and to prevent overfitting, we utilize a combination of random flips and rotation during training. All experiments are conducted on the NIVIDIA RTX A2000 GPU and COLAB V100 GPU.

\subsection{Evaluation Metrics}

Among the metrics used to gauge the performance of diverse FL methods, the F1 score and AUC stand out, capturing the essence of their effectiveness comprehensively. For the 3-category classification, a meticulous evaluation is undertaken by calculating these metrics for each category independently, followed by a macro average to neutralize any influence from class imbalances. Higher scores in both F1 and AUC, like musical notes in perfect harmony, indicate exceptional proficiency in the dermoscopic diagnosis task. In addition to showcasing individual client-specific results, we present the succinct macro average of all four clients, presenting a harmonious comparison that reveals the true prowess of the FL methods.

\subsection{Experimental Results}
We assessed the performance of our personalized meta federated learning framework by conducting a comprehensive comparison with state-of-the-art FL methods, which included FedBN \cite{23}, MetaFed \cite{34}, FedAvg \cite{20}, SiloBN, IDA, FML, and FedProx \cite{30}. 

Our FedMetaMed approach holds substantial promise in addressing key limitations and challenges inherent in traditional federated learning frameworks, particularly concerning scalability issues and privacy concerns. By leveraging meta-learning techniques, the model gains the ability to rapidly adapt to the unique characteristics of diverse local datasets without necessitating the exchange of raw data. This adaptability inherently contributes to mitigating the heterogeneity challenge often faced in federated learning setups. Furthermore, our approach allows for the efficient extraction and utilization of knowledge across multiple tasks, enhancing the model's generalization capability. In terms of scalability, the FedMetaMed framework facilitates the seamless integration of new participants without compromising overall performance. Additionally, by focusing on learning optimal initialization strategies or task-specific adaptations, federated meta-learning minimizes the necessity for extensive communication rounds, thereby alleviating communication overhead and associated scalability issues. The emphasis on learning from a meta-level perspective also aligns with privacy preservation goals, as sensitive information is abstracted into shared meta-knowledge rather than being explicitly transmitted, thereby addressing critical privacy concerns in federated learning scenarios.
% We chose to utilize a setup involving n=4 clients for both training and testing purposes of our proposed model, FedMetaMed. This choice was made after careful consideration of the practical constraints and computational complexities associated with distributed healthcare systems. By selecting a smaller number of clients, we aimed to strike a balance between maintaining a representative dataset while mitigating the challenges posed by heterogeneity in medical data sources. With n=4 clients, we could effectively explore the performance of our Federated Meta-Learning approach in a controlled environment that still encapsulated the diversity of data characteristics present in distributed healthcare systems. While the outcomes presented in this paper are based on n=4 clients, it is plausible that the performance of FedMetaMed could be extrapolated to a larger n clients setting, although this would require addressing additional intricacies and potential bottlenecks associated with an increased number of data contributors. 

Table \ref{table2} presents the results, showing that our method achieved the best performance with an impressive average F1 score of 87\% and average AUC of 90\%. Notably, our framework significantly outperforms the personalized FL method, FedBN, with an average F1 improvement of 9.79\% and an average AUC improvement of 9.86\%.

\begin{figure}[t]
\centering
      \includegraphics[scale=0.3]{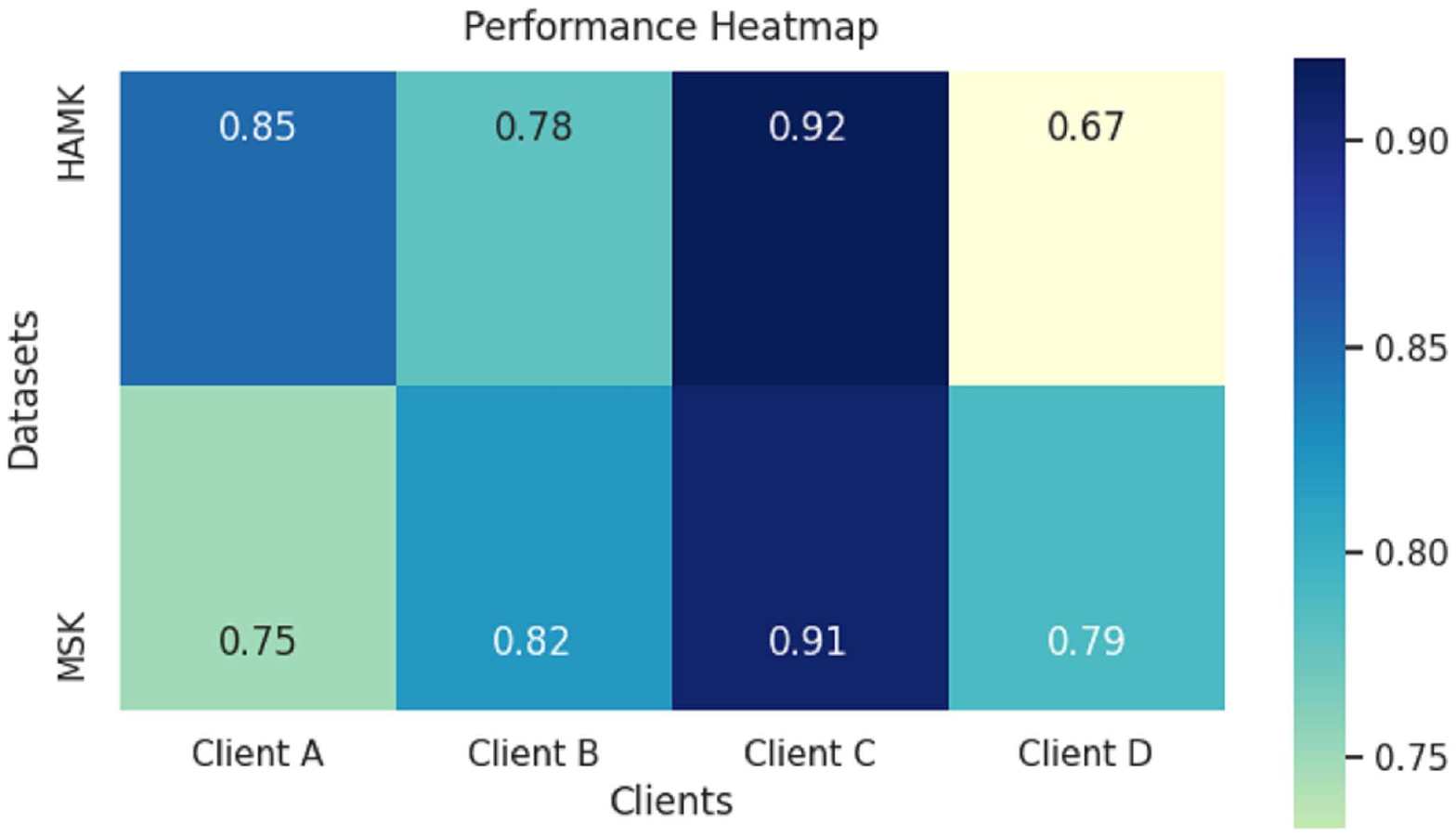}
\caption{Performance comparison across clients for HAMK and MSK dataset.}\label{figureheatmap}
\end{figure}

\begin{figure}[t]
\begin{minipage}{0.22\textwidth}
\centerline{\includegraphics[scale=0.35]{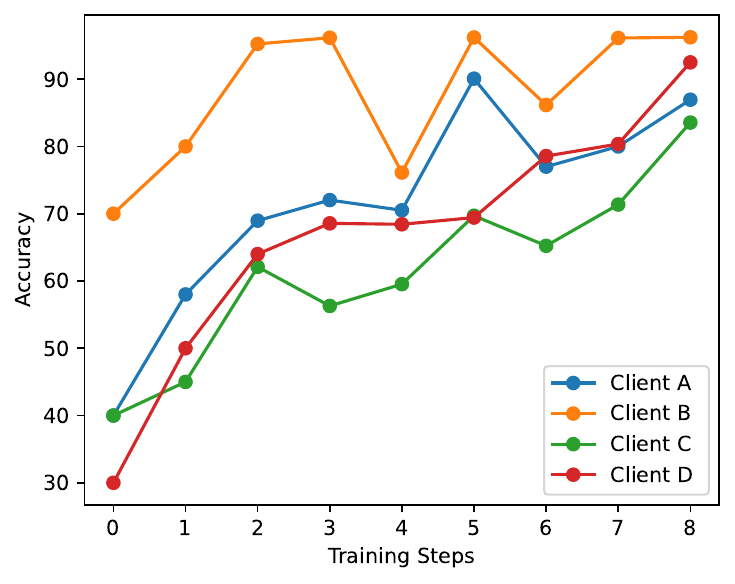}}
\caption{Prediction result in terms of Accuracy with number of iterations.}\label{figacc}
\end{minipage}
\hfill
\begin{minipage}{0.22\textwidth}
\centerline{\includegraphics[scale=0.35]{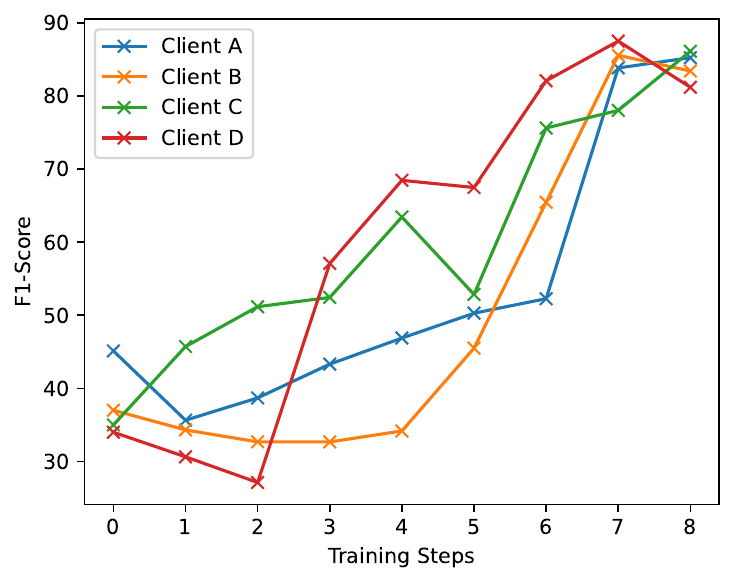}}
\caption{Prediction results in terms of F1-Score with number of iterations.}\label{figf1}
\end{minipage}
\end{figure}

These results confirm that the FedMetaMed CFA and CTO effectively address the limitations observed in FedAvg. Additionally, compared to SiloBN, another collaborative model at each client, our method demonstrates superior performance, with a notable increase of 3.15\% in average F1 and 0.99\% in average AUC. These experimental findings highlight the clear performance advantage of our approach over the state-of-the-art FL methods in real-world medical FL scenarios. Figure \ref{figureheatmap}  shows Dataset wise performance of each client in the form of heatmap.
% \begin{figure}[H]
% \centering
%       \includegraphics[scale=0.40]{ACCURACY.png}
% \caption{Prediction results with number of iterations}\label{fig3}
% \end{figure}
To further establish the effectiveness of CFA (Cumulative Federated Aggregation) and CTO (Collaborative Transfer Optimization) in our personalized federated learning framework, we conducted various experiments. First, we compared the performance of our FedMetaMed framework when removing CTO and CFA individually. Figure \ref{figacc} and \ref{figf1} shows line graph comparsion of all 4-clients in terms of accuracy and F1-Score.

With these changes, we observed that the inclusion of CFA led to an F1 improvement of 10.7\%, and the incorporation of CTO resulted in a remarkable 8.88\% increase in F1. When examining the impact of CFA alone by replacing the parameter aggregation with CFA (without CTO), we found that it contributed to a significant 9.56\% F1 enhancement compared to the standard FedAvg approach.

Furthermore, we evaluated the effect of the same parameter aggregation method for each client (without CFA) and discovered that it outperformed the FedBN approach with a 7\% increase in F1. This improvement can be attributed to the tailored clients utilizing Retrieve-Reciprocate-Refine steps.  Table \ref{table3} shows the generalization results of FedMetaMed compared to existing state-of-the-art techniques on out-of-distribution datasets.
\begin{table}[t]
    \centering
      \caption{Comparison of generalization performance on out-of-distribution data.}
      \scalebox{1.2}
{
    \begin{tabular}{ccc}
   \hline
   % \rowcolor{Blue}
       \textbf{Method} & \textbf{Acc(\%)} & \textbf{F1(\%)} \\ \hline
        FML  &66.8 &  37.05 \\ 
       % \rowcolor{LightBlue}
         IDA  & 62.75 & 40.11 \\ 
        
         SiloBN  & 65.65 & 42.88 \\ 
        % \rowcolor{LightBlue}
         FedBN  & 70.85& 43.76  \\ 
        
        FedAvg  & 66.49 & 38  \\ 
        % \rowcolor{LightBlue}
        FedProx & 67.39 &  38.15\\ 
        MetFed &70 &  54.5\\
        
        \textbf{FedMetaMed} & \textbf{89.51} & \textbf{75.12} \\ \hline
    \end{tabular} }
  
    \label{table3}
\end{table}

\begin{table*}[t]
\caption{The table illustrates a comparison between FedMetaMed and state-of-the-art methods using a real-world medical federated learning dataset. The comparison is based on F1-Score, AUC, Recall Precision metrics across all 4-clients.}
\centering
\scalebox{0.95}
{
\begin{tabular}{lllllllllllllllllllll}
\hline
% \rowcolor{Blue}
\textbf{Method}           & \multicolumn{5}{c}{\textbf{F1(\%)}}  &   \multicolumn{5}{c}{\textbf{AUC(\%)}}&  \multicolumn{5}{c}{\textbf{Pr(\%)}} &\multicolumn{5}{c}{\textbf{Re(\%)}}  \\
\hline
Client                 & \textbf{A}                          & \textbf{B}     & \textbf{C}    & \textbf{D}  & \textbf{Avg}   & \textbf{A}                           & \textbf{B}  & \textbf{C}  & \textbf{D} & \textbf{Avg} & \textbf{A}                           & \textbf{B}  & \textbf{C}  & \textbf{D} & \textbf{Avg} & \textbf{A}                           & \textbf{B}  & \textbf{C}  & \textbf{D} & \textbf{Avg}\\ \hline
                 % \rowcolor{LightBlue}

FML           & 65                        & 72 & 63 & 52& 68  & 83                      & 94 & 83 & 80  & 85 &64&68 &67&74&70&66&73&62&51&64\\ 
IDA         &     55                       &41        &55      & 45   & 51&          81                   & 75     &  78  & 73 &  77 & 72&74&66&77&73&83&76&79&83&81\\ 
 % \rowcolor{LightBlue}
SiloBN          &  50.8          & 63.8        &  53.9     & 61.9  &   55  &     83.1                        & 81.4     &  77   & 80  & 80 & 72&75&76&72&71&80&85&79&84&80 \\ 
FedBN           &   54.6    &     72.1   &  54.3     &62.7 &   65.7   &                    83.0         & 96.3   &   79   &  81 & 88 &73&75&73&63&69&82&91&74&80&82\\ 
 % \rowcolor{LightBlue}
FedAvg            & 57.4                            &48.1        &   56.8    &   44.0    &50 &          81.2                   & 82.7    &   76 & 71  & 79 &74&72&67&73&72&83&85&77&72&75 \\ 
 
FedProx              &  56.7                          &  39.0      &   54.7     &45.5 &  51.3    &          81.7                  &  70.0     &  76   & 74  & 75 &66&72&75&63&70&73&64& 58& 70&69\\ 
MetaFed &59&70&68&82&76&84&82&85&79&80&76&71&69&80&78&82&77.5&83.5&74&76\\
\hline
% \rowcolor{LightBlue}
\textbf{P-CTO} &     67.2                        & 75.2      &     61.4   &  59.4  &  69  &   87.5                           & 89.1    &  83  &  81& 83 &75&77&74&72&74&82&87&75&83 &83 \\ 
\textbf{P-CFA} &  69.1                         & 77.9        &  66.7     &61.9   &   67  &         88.7                    & 97.3     &  84   &  82 & 89  &74&78&77&81&79&83&88&85&80&86\\ 
% \rowcolor{LightBlue}
\textbf{FedMetaMed}         &  82.0                         &  89.7       &   83.7    &  86.6 &    87 &        88.8                     & 97.8    &  85   & 93 & 90 &84&85&83&87&85&88&95&87&84&92\\ \hline
\end{tabular}}
 
    \label{table2}

\end{table*}

% \begin{figure}[H]
% \centering
%       \includegraphics[scale=0.6]{f1score.pdf}
% \caption{Prediction results with number of iterations}\label{fig3}
% \end{figure}

% \begin{figure}[H]
% \centering
%       \includegraphics[scale=0.65]{accuracy.pdf}
% \caption{Prediction results with number of iterations}\label{fig3}
% \end{figure}

\begin{figure}[t]

\centerline{\includegraphics[scale=0.3]{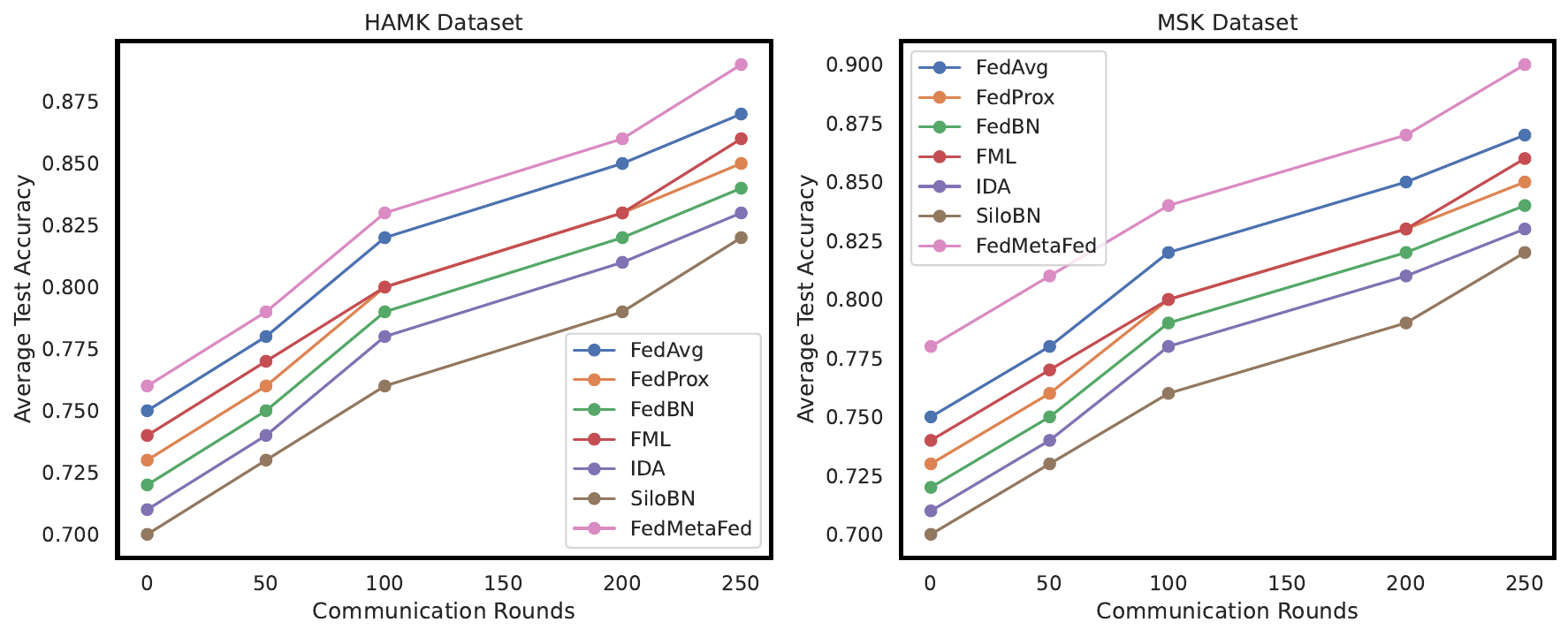}}
\caption{Graphs depicting the test accuracy progression across communication rounds for FedMetaMed, are presented alongside state-of-the-art methods including FedAvg, FedProx, SiloBN, FedBN, and IDA. Each individual subfigure corresponds to distinct cross-datasets: (a) HAMK and (b) MSK.}
\label{figcommround}
\end{figure}

Overall, the ablation experiment affirmed the pivotal role of CFA and CTO in addressing data heterogeneity in real-world personalized medical federated learning and meta-learning scenarios, leading to the superior performance of our personalized federated meta-learning framework.
\paragraph{\normalfont{\textbf{Performance Evaluation w.r.t Communication Rounds}}}
In Figure \ref{figcommround}, the curves depict the mean test accuracy progression over training iterations spanning from 0 to 250 communication rounds. This encompasses the outcomes of all baseline techniques. The findings indicate that, for both HAMK and MSK datasets, FedMetaMed outperforms alternative approaches in terms of average test accuracy and attains quicker convergence.
% \usepackage{graphicx}

\begin{table}[t]
\caption{The table illustrates the outcomes of various ablation studies for FedMetaMed, presenting a comparison of results in relation to metrics such as the AUC  ROC, precision, recall, and F1 score.}
\centering
\scalebox{1.1}
{
\begin{tabular}{llllll}
\hline
% \rowcolor{Blue}
\textbf{Method}         & \textbf{FedMetaMed}         & \textbf{W/FL}          & \textbf{FL}      & \textbf{Meta FL}            \\ \hline
% \textbf{Algo}     & Fed-Meta & Idv-L & FL & Meta-(w/FL)   \\ 
% \rowcolor{LightBlue}
\textbf{AUC (\%)}   & 90.5                    & 84.3                & 87.6             & 89.2                \\ 

\textbf{Pr (\%)} & 87.2                    & 78.9                & 82.4     &       86.8               \\ 
% \rowcolor{LightBlue}
\textbf{Re (\%) }   & 92.1                    & 85.6                & 88.9             &91.3               \\ 
\textbf{F1 (\%) } & 89.5                    & 81.9                & 85.6       &       88.5               \\ \hline
              
\end{tabular}}%
\label{table4}
\end{table}

 % Figure \ref{figcommround} shows test accuracy curves for both datasets HAMK and MSK with respect to communication rounds.

 \paragraph{\normalfont{\textbf{Performance Variation w.r.t Generalization and Heterogenous data}}}
 
To assess the generalization capacity towards out-of-distribution data, we conducted tests on an unseen cohort consisting of 250 samples from a different database. Table~\ref{table4} illustrates the results of various ablation studies conducted with FedMetaMed with and without federated mechanism, meta learning and individual learning algorithms. Our approach exhibits outstanding generalization performance, achieving an F1 score of 89.50\%, Precision of 87.2\%, Recall 92.1\% and an AUC of 90.50\% in this challenging scenario.

\begin{figure}[t]
\centering
      \includegraphics[scale=0.27]{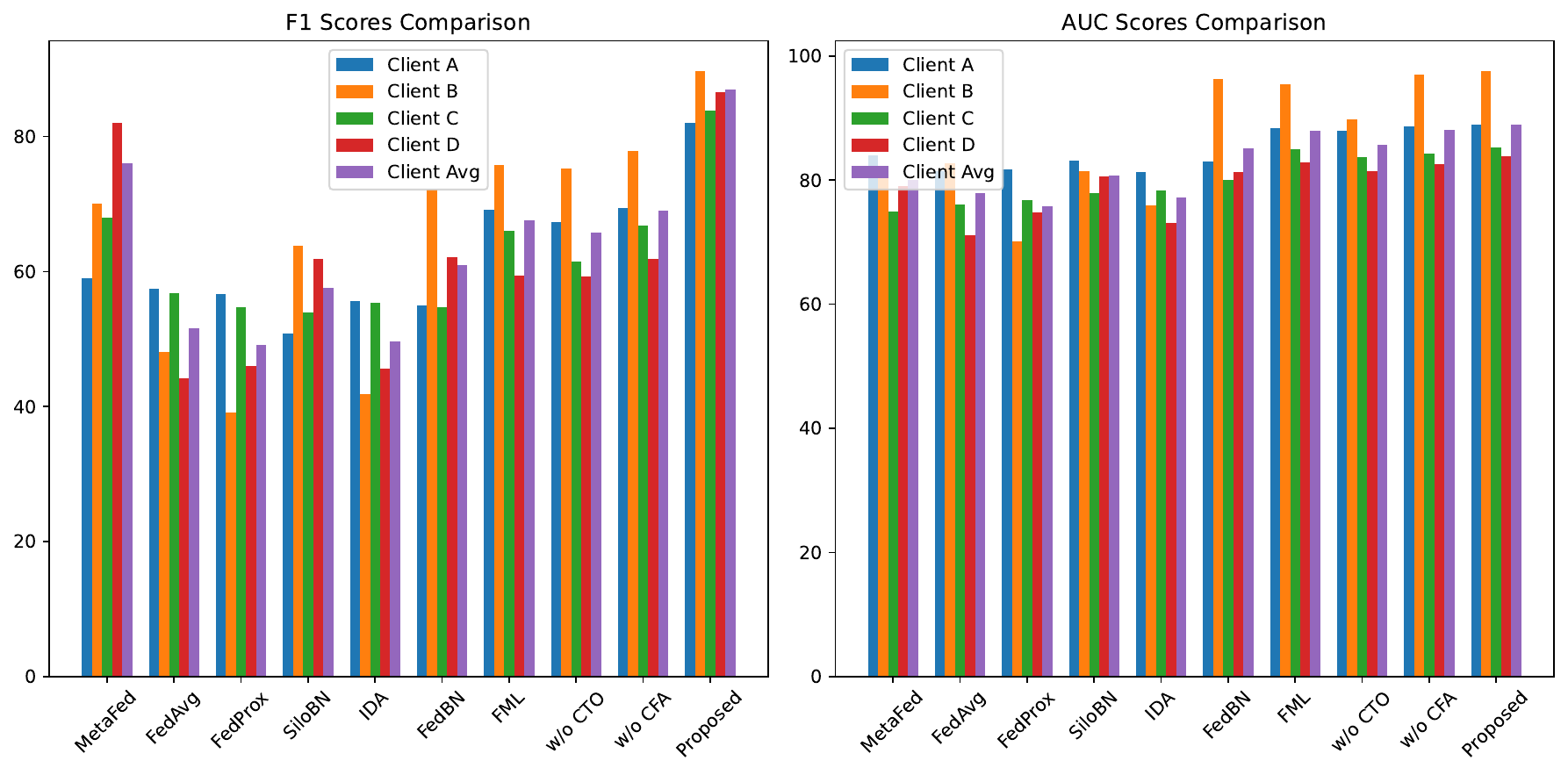}
\caption{Performance comparison across clients of FedMetaMed method from existing techniques.}\label{figcomparison}
\end{figure}
Figure \ref{figcomparison} shows comparative performance bar chart of FedMetaMed with existing state-of-the-art techniques in terms of F1-Score and Accuracy. 
Furthermore, our method showcases superior classification performance when compared to existing algorithms, demonstrating statistically significant improvements of 12.44\%, 12.06\%, 7.33\%, 10.09\%, 6.46\%, and 13.19\% in average F1 score. These competitive experimental outcomes strongly indicate the superiority of our approach in terms of generalization ability. 

\paragraph{\normalfont{\textbf{Analysis of Different Client Numbers}}}
In our proposed FedMetaMed model, we opted to utilize a configuration involving n=4 clients, both for training and testing purposes. This decision was reached after careful consideration of the practical limitations and computational intricacies within distributed healthcare systems. Our aim in selecting this smaller client number was to strike a balance between maintaining dataset representation and managing the complexities arising from diverse medical data sources. By employing n=4 clients, we effectively explored the performance of our Federated Meta-Learning approach in a controlled setting, while still capturing the data diversity seen in distributed healthcare systems.

\begin{figure}[t]
\centering
      \includegraphics[scale=0.4]{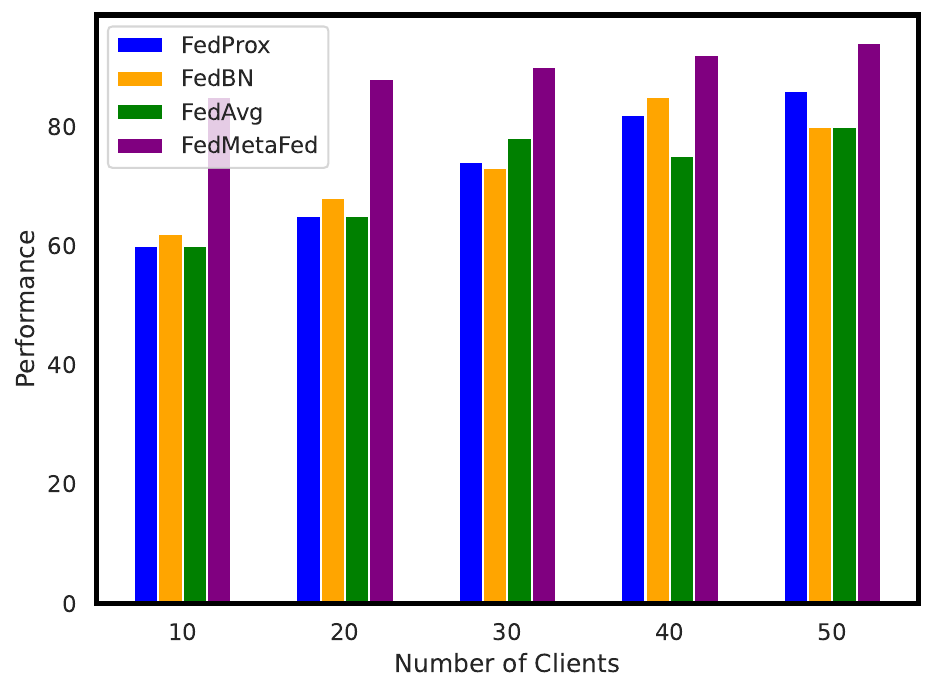}
\caption{Performance of FedMetaMed with State-of-the-Art approaches across different client numbers.}\label{figcounts}
\end{figure}

Although the results presented here are based on the n=4 setup, it is reasonable to extrapolate FedMetaMed's performance to a larger client pool. However, doing so would entail addressing additional complexities and potential bottlenecks tied to an increased number of data contributors. To assess FedMetaMed's capabilities in a broader federated meta learning framework, we extended the client count from 4 to 50. Figure \ref{figcounts} shows performance comparison of FedMetaMed with the existing  state-of-the-art techniques for varying number of clients. The performance of state-of-the-art algorithms initially declines with an increasing number of clients participating in training, owing to heightened divergence. Subsequently, the performance improves due to the decreased difficulty of local tasks with smaller local datasets. These findings suggest that FedMetaMed outperforms alternative techniques in both small- and large-scale federated meta learning systems.

% \begin{longtable}[]
 
%     \begin{tabular}{|c|c|c|c|c|c|c|c|}
%    \hline
%     \textbf{FML} & \textbf{IDA} & \textbf{SiloBN} & \textbf{FedBN} & \textbf{FedAvg} & \textbf{FedProx} & \textbf{Ours} \\ \hline
      
%  A& 69.14& 55.62& 50.83& 54.96& 57.44& 56.70 & 72.00 \\ \hline
%  B&
%  C&
%  D&
%  Avg &
%   A&
%  B&
%  C &
%  D&
%  Avg &

%     \end{tabular}
%     \caption{An analysis of state-of-the-art Federated Learning approaches is conducted to assess their generalization performance on out-of-distribution data}
%     \label{tab:my_label}
% \end{longtable}

% \begin{figure*}[t]

%       \includegraphics[width=6.5cm]{fed_datasets.png}

% \end{figure*}

% \begin{figure*}[t]
% \centering
%       \includegraphics[width=17cm]{enhanced.jpg}

% \end{figure*}

\section{Conclusion}
In this study, we address the challenge of personalized medicine in distributed healthcare using Federated Learning (FL). Our goal is to enhance federated meta-learning for real-world personalized medical applications by introducing modifications at both the server and client levels. At the server, we implement a collaborative Fourier aggregation technique that progressively integrates global insights from low to high-frequency components. On the client side, we employ a three-step Retrieve-Reciprocate-Refine (RRR) approach, which transfers global knowledge to the personalized model without overwriting it, effectively handling heterogeneity issues. Through extensive experiments on a real-world dermoscopic FL dataset, we demonstrate that our framework outperforms existing state-of-the-art methods and generalizes well to an out-of-distribution cohort.
% In this study, we focus on addressing the challenge of Personalized Medication in Distributed Healthcare within the context of existing Federated Learning (FL) methods. Our aim is to enhance the application of federated meta-learning to real-world personalized medical scenarios. To achieve this, we present a personalized framework that introduces modifications both at the server and client levels.

% Specifically, we implement a collaborative Fourier aggregation technique at the server, which progressively combines global insights from low-frequency to high-frequency components. On the client side, instead of outrightly replacing local models, we employ a three-step approach known as Retrieve Reciprocate Refine (RRR). This process effectively transfers centralized knowledge to enhance the personalized model, all the while mitigating issues stemming from heterogeneity. Through extensive experimentation, we validate the superiority of our framework compared to existing state-of-the-art methodologies. This validation is conducted using a real-world dermoscopic Federated Learning dataset, and we also demonstrate its generalizability on an out-of-distribution cohort.

\bibliographystyle{IEEEtran}
\bibliography{aaai}

% \vspace{12pt}
% \color{red}
% IEEE conference templates contain guidance text for composing and formatting conference papers. Please ensure that all template text is removed from your conference paper prior to submission to the conference. Failure to remove the template text from your paper may result in your paper not being published.

\end{document}